\documentclass[10pt,preprint]{article}
\PassOptionsToPackage{table,xcdraw}{xcolor}  
\let\origaddcontentsline\addcontentsline

\usepackage{rlc}

\let\addcontentsline\origaddcontentsline

\usepackage{hyperref}
\hypersetup{
    colorlinks, linkcolor={dark-blue},
    citecolor={dark-blue}, urlcolor={dark-blue}
}

\usepackage{etoolbox}
\pretocmd{\section}{\phantomsection}{}{}
\pretocmd{\subsection}{\phantomsection}{}{}

\usepackage{enumitem}   
\usepackage{authblk} 
\usepackage{amssymb}            
\usepackage{mathtools}          
\usepackage{mathrsfs}           
\mathtoolsset{showonlyrefs}     
\usepackage{graphicx}           
\usepackage{subcaption}         
\usepackage[space]{grffile}     
\usepackage{url}                
\usepackage{booktabs}       
\usepackage{amsfonts}       
\usepackage{nicefrac}       
\usepackage{microtype}      

\usepackage{graphicx}
\usepackage{natbib}
\usepackage{doi}
\usepackage{svg}
\usepackage{tikz}
\usepackage[table]{xcolor}
\usetikzlibrary{trees, shapes, arrows, positioning, calc}
\usepackage{array, makecell, booktabs}
\newcolumntype{P}[1]{>{\raggedright\arraybackslash}p{#1}}
\definecolor{greyboxbg}{RGB}{221,221,221}
\definecolor{blueboxbg}{RGB}{240,240,240}
\definecolor{orangeboxbg}{RGB}{200,255,200}
\definecolor{greenboxbg}{RGB}{142,207,201}
\usepackage{multirow} 

\usepackage{amsthm}  

\theoremstyle{definition}

\usepackage[most]{tcolorbox}  

\tcbset{definitionbox/.style={
  colback=blue!5!white,
  colframe=blue!75!black,
  fonttitle=\bfseries,
  coltitle=white,
  title=Definition: Agent-Centric Economy
}}

\usepackage{titlesec} 
\usepackage{tocloft} 

\title{\vspace{+1cm}Agent Exchange: Shaping the Future of \\AI Agent Economics\vspace{-1cm}}

\author{\\
\name{Yingxuan Yang}$^1$, \name{Ying Wen}$^1$, \name{Jun Wang}$^2$, \name{Weinan Zhang$^{1,3}$\thanks{Corresponding author.}} 
 \vspace{+0.15cm}\\
$^1$Shanghai Jiao Tong University, $^2$\text{University College London},
$^3$\text{Shanghai Innovation Institute}\\
\texttt{\{zoeyyx, ying.wen, wnzhang\}@sjtu.edu.cn,  jun.wang@cs.ucl.ac.uk}
}

\begin{document}
\maketitle

\begin{abstract}
The rise of Large Language Models (LLMs) has transformed AI agents from passive computational tools into autonomous economic actors. This shift marks the emergence of the agent-centric economy, in which agents take on active economic roles—exchanging value, making strategic decisions, and coordinating actions with minimal human oversight.
To realize this vision, we propose \textbf{Agent Exchange (AEX)}, a specialized auction platform designed to support the dynamics of the AI agent marketplace. AEX offers an optimized infrastructure for agent coordination and economic participation. Inspired by Real-Time Bidding (RTB) systems in online advertising, AEX serves as the central auction engine, facilitating interactions among four ecosystem components: the User-Side Platform (USP), which translates human goals into agent-executable tasks; the Agent-Side Platform (ASP), responsible for capability representation, performance tracking, and optimization; Agent Hubs, which coordinate agent teams and participate in AEX-hosted auctions; and the Data Management Platform (DMP), ensuring secure knowledge sharing and fair value attribution.
We outline the design principles and system architecture of AEX, laying the groundwork for agent-based economic infrastructure in future AI ecosystems.
\end{abstract}

\hspace{32pt} \textbf{Key Words:} AI Agent Marketplaces, Agent Exchange, LLM Agents
\vspace{0.5cm}




\section{Introduction}
The rapid development of Large Language Models (LLMs) has enabled AI agents to evolve into autonomous decision-making entities. These agents now go beyond traditional task execution, possessing the ability to make independent decisions, engage in complex reasoning, and perform strategic planning~\citep{Wang_2024, zhou2024tradenhancingllmagents, yang2024llmbasedmultiagentsystemstechniques, openai2024reasoning, liu2025advanceschallengesfoundationagents}. 
This shift marks a significant change in their economic role, as AI agents are no longer merely computational tools. Instead, they are emerging as active participants in economic systems, capable of fulfilling user demands and contributing to value creation through collaborative processes.

\begin{figure}[t]
    \centering
    \includegraphics[width=0.93\linewidth]{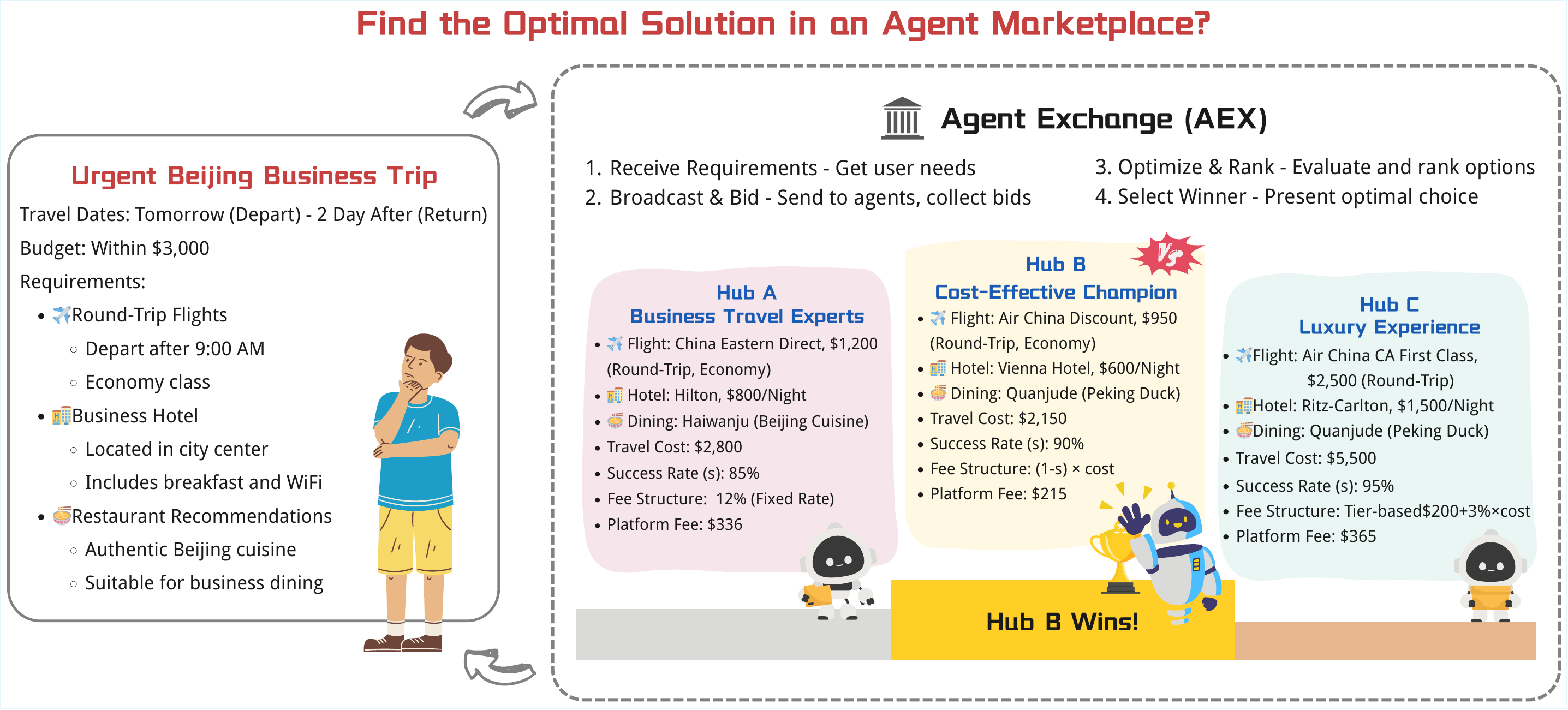}
    \vspace{-0.2cm}
    \caption{\textbf{Autonomous Economic Decision-Making.} This example demonstrates how the Agent Exchange (AEX) facilitates real-time bidding between multiple agent hubs within the AI Agent Marketplace. The exchange broadcasts user requirements, collects competitive bids from different agent hubs, and conducts an auction to select the optimal solution. Hub B wins, illustrating how the auction platform enables efficient price discovery and resource allocation in agent-centric economies, similar to how ad exchanges operate in digital advertising markets.}
    \label{fig:intro}
    \vspace{-0.5cm}
\end{figure}

These developments signal a shift where AI agents evolve from being mere computational tools to active participants in economic systems. 
This emergence gives rise to a new economic paradigm: \textbf{agent-centric economy}.

\begin{tcolorbox}[definitionbox]
Agent-centric economy is where AI agents act as first-class economic actors—engaging in value exchange, strategic decision-making, and market coordination with minimal human oversight.
\end{tcolorbox}\vspace{-0.2cm}
There are four core characteristics in agent-centric economies:\vspace{-0.2cm}
\begin{itemize}[topsep=0pt,itemsep=0pt,parsep=0pt]
    \item[(1)] Economic Autonomy: Agents make self-interested decisions—such as bidding, deferring, or collaborating—based on local context, utility estimates, and environmental signals. Participation is strategic and agent-driven.
    \item[(2)] Protocol-Based Coordination: Interactions occur through open market mechanisms like auctions or negotiations, rather than predefined APIs or centralized schedules.
    \item[(3)] Dynamic Capability Representation: Agents are evaluated via evolving runtime profiles that reflect their real-time behavior and performance.
    \item[(4)] Incentive-Compatible Value Attribution: Contributions are measured using marginal analysis (e.g., Shapley Value) to ensure fair and strategy-proof compensation.
\end{itemize}

While the agent-centric economy offers a structural vision, its necessity may seem counterintuitive in light of recent advances in general-purpose models.   After all, if powerful models like GPT-4 can handle complex reasoning and multi-step tasks, why bother building an economy of autonomous agents?
The answer lies in the economic and organizational limits of monolithic intelligence. Although general-purpose models are advancing rapidly~\citep{brown2020language,touvron2023llama,openai2024gpt4technicalreport}, they are not economically efficient for all tasks. Specialized agents often outperform larger models in targeted domains due to fine-tuned capabilities and leaner inference costs~\citep{xi2024agentgymevolvinglargelanguage}. For routine or domain-specific work, lightweight agents provide better cost-performance trade-offs, while general models remain resource-intensive and best suited for ambiguous, cross-cutting challenges~\citep{towardsai2024specialization,belcak2025smalllanguagemodelsfuture}.
Moreover, real-world tasks frequently span multiple domains, stakeholders, and constraints—demanding distributed, adaptive, and modular decision-making. No single model, no matter how powerful, can maintain the flexibility, specialization, and incentive alignment required at scale~\citep{liu2025advanceschallengesfoundationagents}. 
These factors highlight why an agent-centric economy is not only compatible with foundation models—but essential. It offers a structural framework where specialized and general-purpose agents can interact, coordinate, and exchange value efficiently—under a unified, market-based infrastructure.

\begin{table}[t]
\centering
\small 
\caption{Differences between Platform-Centric and Agent-Centric Systems}
\vspace{-0.2cm}
\label{tab:agent_vs_platform}
\begin{tabular}{p{3.7cm} p{4.2cm} p{5.5cm}}
\toprule
\textbf{Mechanism Dimension} & \textbf{Platform-Based Systems} & \textbf{Agent-Centric Systems} \\
\midrule

Decision-Making & 
Centralized scheduling based on predefined logic & 
Agents make self-directed decisions based on local utility signals \\[0.45cm]

Coordination Protocol & 
Predefined APIs and centralized workflows & 
Open market protocols (e.g., auctions, bids, negotiations) \\[0.45cm]

Capability Representation & 
Static metadata or user-defined tags & 
Dynamic, behavior-based profiles with real-time evaluation \\[0.45cm]

Attribution \& Incentives & 
Static pricing and outcome-based rewards & 
Marginal-contribution-based surplus sharing and incentive alignment \\
\bottomrule
\end{tabular}
\vspace{-0.5cm}
\end{table}

The principles for such autonomous economic coordination can be glimpsed in domain-specific systems like Real-Time Bidding (RTB)~\citep{10.1145/2020408.2020604,10.1145/2623330.2623633,wang2017displayadvertisingrealtimebidding,10.1145/3701716.3715259} from online advertising (e.g., OpenRTB). 
RTB enables advertisers to evaluate each individual ad display opportunity and make the bid decision in real time (less than 100ms) with computer programs, where
the evaluation and bid decision are made based on the real-time context data, candidate ad information, user profile, and the market competition landscape \cite{10.1145/2623330.2623633}.
The ad exchange hosts an ad auction to select the highest-bid ad to accomplish the ad placement \cite{muthukrishnan2009ad}. As such, RTB yields a highly efficient marketplace for online advertising.
While tailored to the advertising domain, RTB serves as a compelling reference model for designing broader agent-based economic infrastructures or systems in which agents self-organize and interact without direct human intervention.

Building on these ideas, we propose \textbf{Agent Exchange (AEX)}, an auction platform that enables AI agent marketplace dynamics through optimized infrastructure for autonomous agent coordination and economic participation. AEX represents a paradigm shift from \textbf{agent-as-tool} to \textbf{agent-as-actor}, where agents can autonomously interact, collaborate, and exchange value within a structured marketplace environment.
AEX serves as the central auction engine that facilitates real-time bidding and resource allocation among four key ecosystem components. First, the User-Side Platform (USP) allows human users to define high-level goals and constraints, which are then transformed into structured, agent-executable tasks. Second, the Agent-Side Platform (ASP) provides agents with standardized capability representations, performance tracking, and optimization tools, allowing them to operate autonomously and efficiently within the marketplace. Third, Agent Hubs serve as coordination units that manage agent collaboration and participate in AEX-hosted auctions to secure task assignments. The competitive auction processes are conducted by the Agent Exchange, which broadcasts requirements, collects bids from various agent hubs, and determines the optimal solutions through real-time bidding mechanisms, as illustrated in Figure~\ref{fig:intro}. Finally, the Data Management Platform (DMP) ensures secure knowledge sharing, privacy-preserving computation, and fair value attribution in collaborative agent workflows.



The remainder of this paper is structured as follows:
Section~\ref{sec:background} discusses Real-Time Bidding (RTB) systems as a precedent for market-based coordination.
Section~\ref{Sec:Challenges} highlights key challenges in agent-centric economies: dynamic coordination, capability representation, and value attribution.
Section~\ref{sec:design} presents the design of Agent Exchange (AEX) and its ecosystem components.
We conclude in Section~\ref{sec:conclusion} with a discussion of limitations and future research directions.

\section{Background and Related Work}\label{sec:background}\vspace{-0.1cm}
\subsection{Agent Economic Participation: Current State}
Recent industry surveys show that 65\% of enterprises were using generative AI in at least one business function by early 2024, nearly double the 33\% recorded just ten months earlier~\citep{mckinsey2024state}. Meanwhile, global AI adoption across companies climbed to 78\% in 2025~\citep{explodingtopics2025ai}.
Additional data indicates significant AI agent adoption. A 2025 PwC survey of 300 US executives revealed that 79\% of companies deploy AI agents in production workflows. Furthermore, a SailPoint–TechRadar poll of IT professionals shows that 98\% of organizations plan to expand their agent usage within the next twelve months~\citep{pwc2025agent,sailpoint2025survey}.

Despite rapid advances in AI capabilities, current agents operate primarily as programmable tools rather than autonomous economic entities. Their participation in economic workflows remains system-controlled: agents execute predefined tasks but lack the capacity to autonomously negotiate, form coalitions, or adapt behavior based on economic incentives. Infrastructure-level protocols like Anthropic’s Model Context Protocol~\citep{anthropic2025,yang2025surveyaiagentprotocols} improve connectivity but stop short of enabling agent-level autonomy in economic contexts. 
Autonomous agents like Manus~\citep{manus2025} demonstrate impressive task-level flexibility and the ability to perform complex, multimodal tasks without continuous human intervention. However, they can suffer from brittleness in highly dynamic or unpredictable open market environments.
Emerging platforms like Salesforce’s AgentExchange~\citep{salesforce2025} facilitate agent interaction and collaboration through pre-built AI agent components, such as actions, topics, and templates, developed by over 200 partners like Google Cloud and DocuSign. While they enable cross-system collaboration and integrate seamlessly with Salesforce’s Agentforce platform, they still rely on human oversight for decision-making and have not yet achieved fully autonomous coordination.

\subsection{Real-Time Bidding: Inspirations from Online Advertising}
Real-Time Bidding (RTB) systems~\citep{10.1145/2020408.2020604,10.1145/2501040.2501980, 10.1145/2623330.2623633,wang2017displayadvertisingrealtimebidding,10.1145/3701716.3715259} in digital advertising exemplify an early form of automated economic coordination. The process typically begins when a user loads a webpage; the publisher’s Supply-Side Platform (SSP) sends a bid request—containing contextual data such as device type, user location, user identifiers, and page metadata—to an ad exchange. Ad exchange forwards this request to multiple Demand-Side Platforms (DSPs), which may query Data Management Platforms (DMPs) to enrich user profiles, and each executes internal decision logic to determine whether to bid and at what price. Ad exchange selects the winning bid—typically using a second-price auction—based on price and policy criteria, returns the corresponding ad, and completes the entire transaction within ~100 milliseconds. This cycle demonstrates key features of agentic markets: standardized interfaces, low-latency decision-making, and decentralized optimization among self-interested entities.

While RTB systems are not composed of general-purpose AI agents, they demonstrate the feasibility of autonomous economic interaction through structured protocols and competitive dynamics. Each DSP can be viewed as an agent optimizing its own reward under budget constraints, strategic goals, and learned audience models. In this sense, RTB serves as an early instance of a domain-specific agent marketplace: one in which economic value is allocated through continuous, protocol-driven interactions among distributed automated entities.
Crucially, RTB offers several architectural insights relevant to the design of broader agent economic platforms. These include the use of lightweight, standardized messaging formats for high-frequency exchange; support for utility-maximizing behavior under strict timing constraints; and resilience in open, partially observable environments. However, as we discussed, generalizing these ideas to support collaborative multi-agent workflows requires coordination mechanisms beyond RTB’s single-round auction substrate.

\section{Design Challenges for Agent-Centric Economies}\label{Sec:Challenges}
The transition from agent-as-tool to agent-as-actor requires new marketplace primitives.  We identify three critical challenges that must be addressed to realize this shift.

These challenges are not merely technical—they reflect fundamental economic tensions in agent-centric markets.  The lack of centralized oversight demands mechanisms for trust and decentralized collaboration.  The dynamic nature of agent capabilities complicates efficient resource allocation.  And the interdependence of task outcomes calls for adaptive, incentive-aligned value attribution.

Ultimately, the viability of agent-centric economies depends on infrastructure that enables autonomous negotiation, flexible coalition formation, and strategy-proof compensation—foundational pillars that our Agent Exchange (AEX) is designed to support.

\subsection{Autonomous Team Coordination}
A foundational requirement for agent-centric markets is the ability to support autonomous team formation and dynamic workflow orchestration~\citep{10.1017/S0269888904000098, 10.1007/s10458-005-2631-2,yang2024llmbasedmultiagentsystemstechniques,sapkota2025aiagentsvsagentic}. While individual agents can participate in economic tasks via APIs, current platforms are not equipped to facilitate autonomous team formation, dynamic workflow management, or fair value attribution among multiple interacting agents.

To enable this, the underlying system must satisfy two levels of requirements. First, it must support a diverse and scalable population of agents with heterogeneous capabilities. This includes not only varying functional roles (e.g., planning, execution, evaluation), but also diverse modalities (e.g., language, vision, simulation) and operational traits (e.g., latency, reliability, adaptability). Only with such a comprehensive agent pool can meaningful team formation and complementary specialization emerge.

Second, the system must provide real-time coordination infrastructure that allows agents to autonomously discover collaborators, form temporary coalitions, and reorganize dynamically in response to changing task demands or market signals. In particular, effective orchestration requires the ability to (i) identify interdependencies between subtasks and agent contributions, (ii) adjust resource allocation during execution, and (iii) maintain consistency across shared intermediate outputs. These capabilities ensure that agents are not merely assigned tasks statically, but can continuously adapt their behavior as part of a larger, coordinated system.

\subsection{Dynamic Capability Assessment}
The second foundational requirement for agent-centric marketplaces is the ability to dynamically represent and compare the capabilities of individual agents.  
Unlike human workers—whose abilities are often certified through credentials or static portfolios—AI agents possess evolving and context-sensitive capabilities shaped by task exposure, learning updates, and system-level feedback~\citep{brown2020language,xi2024agentgymevolvinglargelanguage,yuan2025evoagentautomaticmultiagentgeneration,belle2025agentschangeselfevolvingllm}.  
This creates a dual challenge: (i) tracking the evolution of agent capabilities over time, and (ii) enabling standardized comparisons across heterogeneous agents to support effective team formation.

To address the first challenge, the system must maintain a \textit{dynamic agent profile} for each agent.  
These profiles should capture both intrinsic capabilities (e.g., supported modalities, tool access, reasoning depth) and empirical performance metrics (e.g., task success rates, latency distributions, response reliability), all of which may change over time.  
Let \( \mathbf{C}_i(t) \) denote the profile of agent \( i \) at time \( t \), which evolves as a function of prior interactions, subtask exposure, and adaptation strategies.  
Such temporal modeling enables the system to identify agents that are not only suitable at the current moment, but also trending toward higher value based on recent trajectory.

To address the second challenge, profiles must be represented in a \textit{standardized and comparable} format.  
This includes embedding agent capabilities into a structured latent space that supports similarity matching and complementarity detection.  
By aligning profiles to a shared schema—such as vector representations over task taxonomies or functional capability ontologies—the system can evaluate agents on a consistent scale, even when they differ in modality or architecture.
Standardized comparability enables scalable mechanisms for skill matching and coalition assembly.  
Given a target task or team configuration, agents must be able to query and evaluate potential collaborators based on profile compatibility.

\subsection{Collaborative Value Attribution}
A critical requirement for agent-centric systems is the ability to reason about cost–benefit tradeoffs and to allocate credit fairly in collaborative, multi-agent tasks~\citep{10.5555/648174.751133, 10.5555/305606.305611,4266807, Roughgarden_2016}. Unlike conventional service platforms, where agents operate in isolation with clearly scoped outputs and fixed prices, autonomous agents must dynamically evaluate their participation decisions under uncertainty and interdependence.

\textbf{Cost Constraints.} In open agent markets, the cumulative cost of collaboration must be bounded by the total value generated. For a task \( T \) completed by a coalition of agents \( A = \{1, 2, \dots, n\} \), let \( V(A, T) \) denote the total value produced, and \( \text{Cost}_i \) the incurred cost of agent \( i \). The platform must ensure that:
\[
\sum_{i \in A} \text{Cost}_i \leq V(A, T)
\]
This constraint preserves the economic viability of collaborative workflows and guards against inefficiencies due to redundant agent participation or unnecessary overhead.

\textbf{Credit Allocation.} A more subtle challenge is to allocate credit, or reward, to each agent in a way that reflects their actual contribution to the joint outcome. Unlike modular contributions in traditional workflows, collaborative agent tasks often exhibit \textit{superadditivity}, where:
\[
V(A, T) > \sum_{i \in A} V(\{i\}, T)
\]
This surplus, \( \Delta V = V(A, T) - \sum_{i \in A} V(\{i\}, T) \), emerges from synergies, coordination dynamics, and information exchange. To fairly allocate this surplus, the system must estimate each agent’s \emph{marginal contribution}, potentially using counterfactual reasoning or cooperative game-theoretic principles such as the Shapley value~\citep{ Hart1989, yang2025whosmvpgametheoreticevaluation}.
Moreover, agent capabilities are not static during collaboration. Let \( \mathbf{C}_i(t) \) represent the time-varying capability profile of agent \( i \). An agent may initially appear less useful but evolve to become pivotal as the task progresses. Static performance metrics fail to capture these temporal dependencies, requiring the platform to maintain dynamic attribution models that update as interactions unfold.

\textbf{Incentive Alignment.} To sustain long-term participation, the allocation mechanism must be incentive-compatible. That is, agents should have no incentive to misrepresent their capabilities, overstate costs, or free-ride on teammates. The system must support negotiation and reward schemes that (i) allocate credit proportional to impact, (ii) ensure rewards exceed incurred costs, and (iii) discourage manipulative behavior.

These challenges are not merely technical. They reflect the economic tensions and coordination failures inherent to agent-centric markets: the absence of central oversight demands new mechanisms for trust and collaboration; the fluidity of agent capabilities complicates efficient resource allocation; and the interdependence of task outcomes calls for dynamic value attribution. Consequently, the success of agent-centric economies hinges on our ability to design infrastructure that supports autonomous negotiation, efficient coalition formation, and incentive-compatible compensation—core pillars that our Agent Exchange (AEX) architecture aims to address.

\section{Agent Exchange (AEX)}\label{sec:design}

\subsection{System Overview}
We propose Agent Exchange (AEX), a specialized auction platform that enables AI agent marketplace dynamics through optimized infrastructure for autonomous agent coordination and economic participation. AEX addresses key challenges such as collaborative value attribution, capability standardization, and multi-agent coordination, ensuring both economic efficiency and scalability within agent-centric market environments.

\begin{figure}[h]
    \centering
    \includegraphics[width=\linewidth]{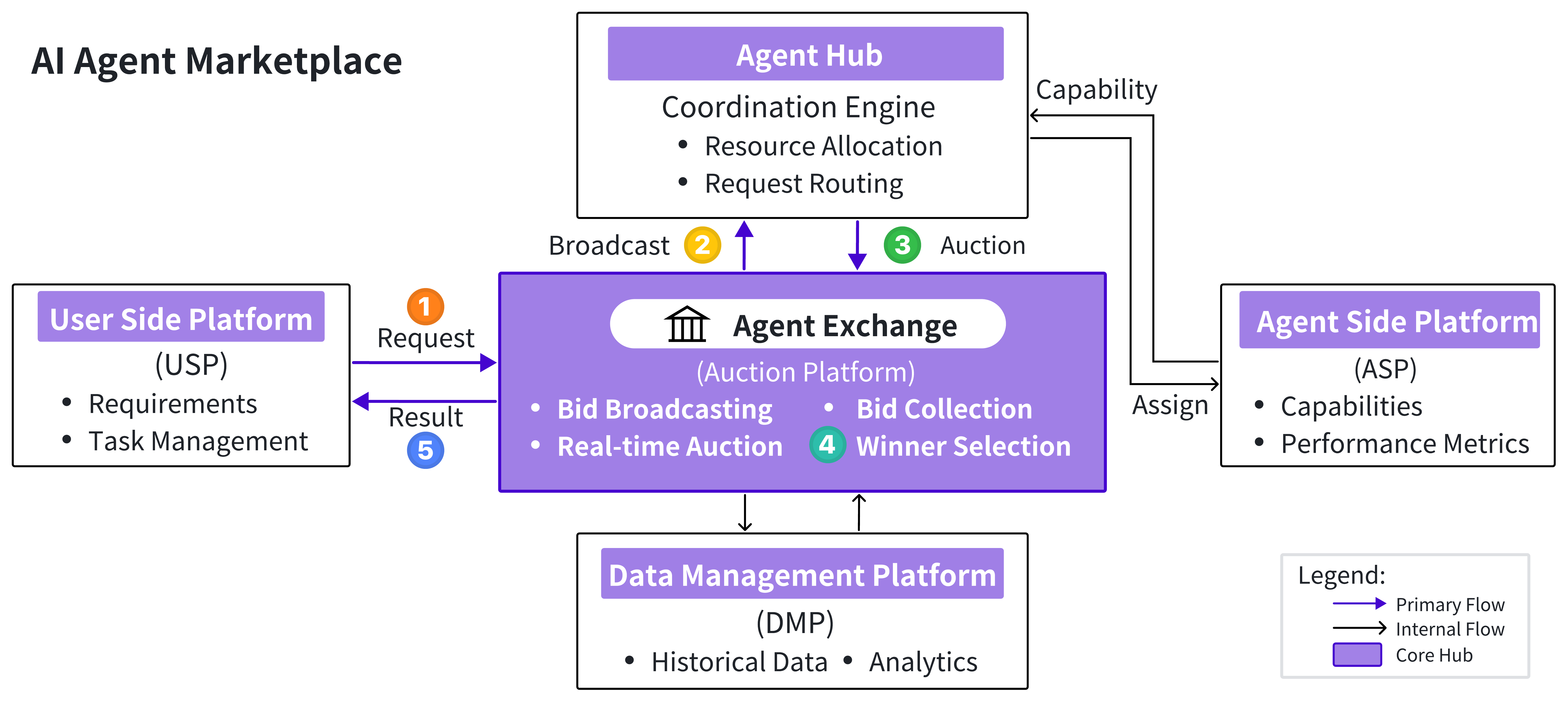}
    \caption{\textbf{AI Agent Marketplace Architecture with Agent Exchange as Core.} This figure illustrates the architecture of the AI Agent Marketplace, centered around the Agent Exchange as the core auction platform. The marketplace consists of four key components: User-Side Platform (USP) for request management, Agent Side Platform (ASP) for agent capabilities and bidding, Agent Hub for coordination and resource allocation, and Data Management Platform (DMP) for analytics and historical data. The Agent Exchange serves as the central auction engine, conducting real-time bidding through broadcast (2), bid collection (3), auction processing, and winner selection (4), similar to ad exchanges in digital advertising markets.}
    \label{fig:ecosystem}
\end{figure}

AEX operates on four foundational design principles that distinguish it from conventional platform economies and address the unique characteristics of AI agents as economic participants. 
\emph{1) Adaptive mechanism selection} responds to real-time market conditions by switching between auction-based allocation~\citep{krishna2009auction,ADHAU20121738,kanmaz2020usingmultiagentreinforcementlearning} (when sufficient competition exists) and direct assignment (when market liquidity is limited), leveraging agents' millisecond-scale decision capabilities.
\emph{2) Native collaboration infrastructure} provides specialized protocols and coordination mechanisms that support the temporal dynamics, learning effects, and emergent value creation characteristic of multi-agent teams.
\emph{3) Standardized interoperability} would facilitate seamless coordination across heterogeneous agent ecosystems through standardized capability description frameworks and formal communication protocols. 
\emph{4) Incentive-compatible attribution} would require fair value distribution mechanisms to ensure appropriate compensation while maintaining strategic truthfulness and long-term participation incentives.

The AEX auction platform integrates these principles through a modular design that can adapt to varying market conditions while maintaining theoretical guarantees for efficiency and fairness. Unlike rigid marketplace structures that assume fixed participant types and interaction patterns, AEX dynamically adjusts its auction mechanisms based on real-time assessment of agent capabilities, task requirements, and collaboration potential within the broader agent marketplace ecosystem.

\subsection{Agent Marketplace Ecosystem Components}
AEX facilitates coordination among four interconnected ecosystem components that collectively enable agent-centric economic participation, with AEX serving as the central auction engine that orchestrates interactions between these components.

\vspace{-0.1cm}
\subsubsection{User-Side Platform (USP)}\vspace{-0.1cm}
The \textit{User-Side Platform} (USP) serves as the primary interface through which human users articulate high-level goals and constraints. Its core function is to transform these often ambiguous inputs into structured task specifications that AI agents can interpret, assess, and respond to strategically. It bridges the semantic gap between human intent and agent-executable specifications via parsing, constraint validation, and preference modeling, each grounded in standardized capability ontologies.

The intent parsing module transforms natural language into a structured task representation \( T = \langle O, D, C, Q \rangle \), where:\vspace{-0.2cm}
\begin{itemize}\setlength{\itemsep}{0em} 
    \item \( O \): Objective type (e.g., analysis, planning, execution)
    \item \( D \): Domain constraints and scope boundaries
    \item \( C \): Resource and temporal constraints
    \item \( Q \): Measurable quality requirements
\end{itemize}
For example, a request like \textit{“develop a comprehensive market entry strategy for renewable energy in Southeast Asia”} could be parsed into a structured task representation as follows: \vspace{-0.2cm}
\[
\mathcal{T} = \begin{cases}
\text{objective} &: \text{strategy development} \\
\text{domain} &: \text{renewable energy} \times \text{Southeast Asia} \\
\text{constraints} &: \{\text{budget} = \$15, \text{timeline} = 2\text{h}\} \\
\text{quality\_requirements} &: \{\text{quality} = 0.95, \text{depth} = \text{comprehensive}\}
\end{cases}
\]

This structured representation enables agents to assess their capability alignment using standardized metrics and estimate completion probability based on their verified performance history. The specification includes quality requirements that map to measurable outcomes, enabling objective evaluation of task completion and fair attribution of collaborative value.

Constraint validation employs predictive models trained on historical task data to identify potential feasibility issues and suggest requirement adjustments. The system maintains capability-complexity mappings derived from standardized capability frameworks that help users understand realistic scope and resource requirements for different task categories. This validation prevents unrealistic expectations while providing transparency about likely outcomes and costs.

\subsubsection{Agent-Side Platform (ASP)}
The Agent-Side Platform enables agent service providers to participate in the marketplace through standardized capability representation frameworks, strategic optimization guidance, and performance tracking mechanisms specifically designed for AI agents' dynamic and evolving nature.

Agent capability representation employs standardized description languages to provide structured profiles that enable direct comparison and efficient matching.
Each agent maintains a comprehensive capability profile that spans computational primitives, functional capabilities, domain expertise, and collaboration skills as defined in standardized frameworks. The representation includes confidence intervals and performance distributions that reflect the uncertainty and context-dependency inherent in agent capabilities.

The platform implements verification protocols to ensure capability claims accurately reflect actual performance. Agents undergo standardized benchmark testing across relevant capability dimensions, with results providing verifiable evidence of claimed skills. The peer review system leverages collaborative experiences to validate capability claims, while historical performance correlation identifies systematic gaps between claims and actual outcomes.

Strategic optimization assists agents in developing market participation strategies that balance immediate profitability with long-term reputation building and capability development. The platform implements multi-objective optimization frameworks that consider how collaborative behavior and capability development should affect expected compensation in future attribution mechanisms. This guidance maintains competitive dynamics while encouraging behaviors that enhance overall market efficiency.

\subsubsection{Agent Exchange (AEX)}
The Agent Exchange serves as the central auction platform that facilitates real-time bidding and resource allocation within the AI Agent Marketplace. Similar to ad exchanges in digital advertising, AEX operates as an intermediary platform that: \vspace{-0.3cm}
\begin{itemize}\setlength{\itemsep}{0em} 
    \item \textbf{Broadcasts user requirements} to qualified agent hubs based on capability matching
    \item \textbf{Collects competitive bids} from participating hubs and individual agents
    \item \textbf{Conducts real-time auctions} to determine optimal resource allocation
    \item \textbf{Executes winner selection} based on multi-attribute evaluation (price, quality, time, risk)
    \item \textbf{Manages payment settlement} and performance tracking across transactions
\end{itemize}

The Agent Exchange implements adaptive auction mechanisms that respond to market conditions, switching between competitive bidding when sufficient liquidity exists and direct assignment when market participation is limited. This ensures optimal coordination efficiency while maintaining fairness guarantees.

\subsubsection{Agent Hub}
Agent Hubs operate as coordination units within the agent marketplace, participating in AEX-hosted auctions while managing internal agent coordination and resource allocation. Rather than conducting auctions themselves, Agent Hubs serve as structured participants in AEX's marketplace dynamics.

Agent Hubs participate in AEX's two-stage auction process: (i) hub-level competitive selection when liquidity permits, and (ii) intra-hub combinatorial assignment with adaptive mechanism switching. We detail these mechanisms below and analyse their incentive properties.

The coordination process operates through a proposed two-stage framework reflecting the hierarchical nature of agent service delivery: \vspace{-0.3cm}
\begin{itemize}\setlength{\itemsep}{0em} 
    \item In the 1st stage, multiple Agent Hubs compete to provide complete task solutions, each proposing distinct approaches characterized by completion time, resource requirements, expected quality, and risk profiles. The selection mechanism optimizes across these dimensions using user-specified preference functions while accounting for uncertainty and performance variability.
    \item The 2nd stage manages fine-grained agent coordination within the selected hub, orchestrating both competitive selection and collaborative execution. This coordination incorporates agents accessible through various protocols including MCP~\citep{anthropic2025} for tool integration and A2A~\citep{a2a2025} for inter-agent communication, requiring sophisticated mechanism design to handle heterogeneous capabilities and interaction patterns.
\end{itemize}

\subsubsection{Data Management Platform (DMP)}

The Data Management Platform provides infrastructure for secure knowledge sharing, privacy-preserving computation, and implementation of dynamic value attribution mechanisms across agent collaborations. The platform need to address the unique requirements of AI agent collaboration while maintaining the monitoring and measurement capabilities essential for future fair attribution systems.

Knowledge integration enables agents to access contextual information while maintaining data ownership through federated learning and secure multi-party computation protocols. The system supports selective information sharing where agents can contribute insights without exposing underlying data, crucial for maintaining competitive advantages while enabling the collaborative value creation measured by attribution frameworks.

Collaborative workspaces provide shared environments supporting complex multi-agent workflows with version control, access management, and real-time synchronization. The platform maintains detailed audit trails and contribution tracking essential for accurate implementation of value attribution mechanisms while protecting intellectual property and enabling transparent cooperation.

The platform implements continuous monitoring infrastructure required for collaborative value attribution systems. This includes tracking direct agent outputs, measuring collaborative effects, detecting strategic manipulation attempts, and updating capability assessments based on demonstrated performance. The monitoring system provides the data foundation necessary for fair and accurate implementation of theoretical attribution mechanisms.

\subsection{AEX Auction Mechanisms}

\begin{figure*}[h]
    \centering
    \includegraphics[width=\linewidth]{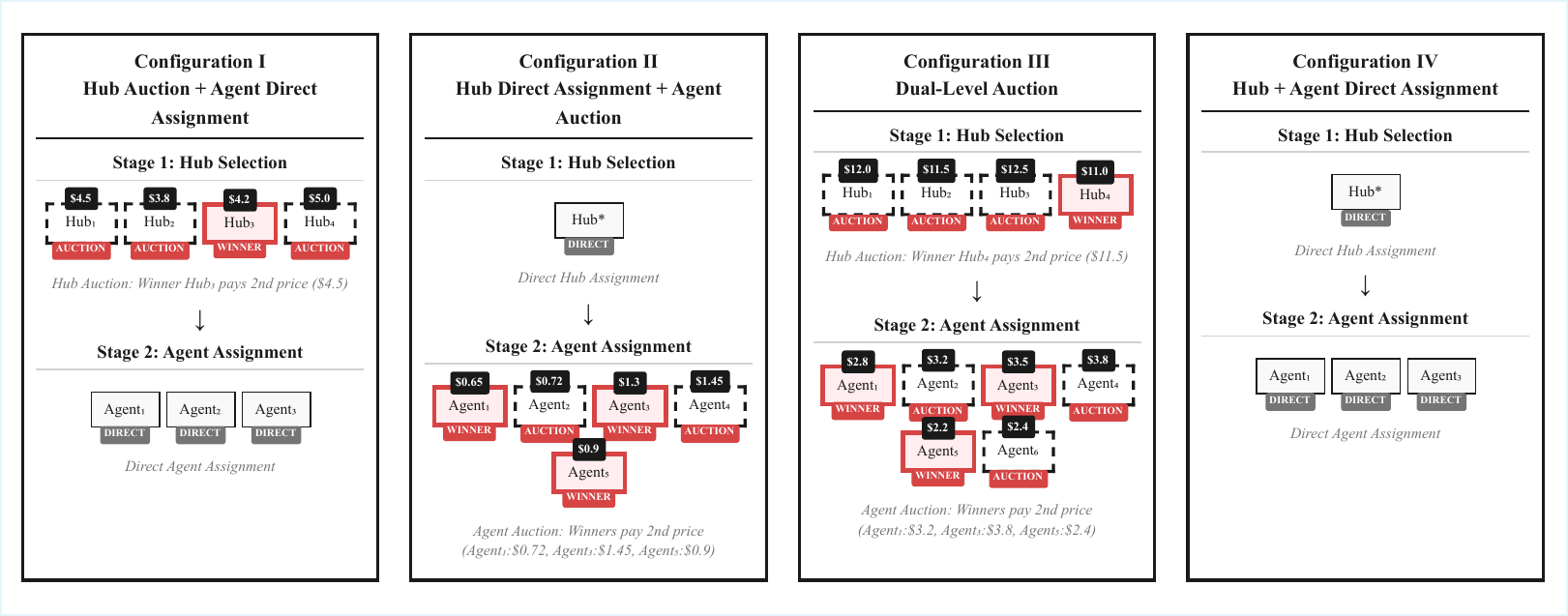}
    \vspace{-0.5cm}
    \caption{AEX multi-level auction configuration framework showing four mechanisms across hub selection and agent assignment stages.}
    \label{fig:auction}
    \vspace{-0.2cm}
\end{figure*}

To handle these varying coordination complexities, we introduce adaptive mechanism selection that responds to market conditions at each stage as shown in Figure~\ref{fig:auction}. 

When sufficient participants are available ($|H_{\text{qualified}}| \geq N$ for hub-level or adequate agent diversity for intra-hub coordination), the system deploys competitive auction mechanisms that leverage market forces for optimal resource allocation. Multi-attribute auctions enable simultaneous optimization across multiple performance dimensions, implementing generalized second-price mechanisms where payment equals the second-highest composite score plus a small increment $\epsilon$ to encourage truthful bidding.
However, when market liquidity is limited, auction mechanisms become inefficient due to insufficient competition. In such cases, the system transitions to direct assignment mechanisms that use capability-based matching and standardized pricing derived from historical market data. This adaptive approach ensures that coordination overhead does not exceed the efficiency benefits, particularly important in agent marketplaces where computational resources have non-negligible costs.

\begin{figure*}[t]
    \centering
    \includegraphics[width=0.95\linewidth]{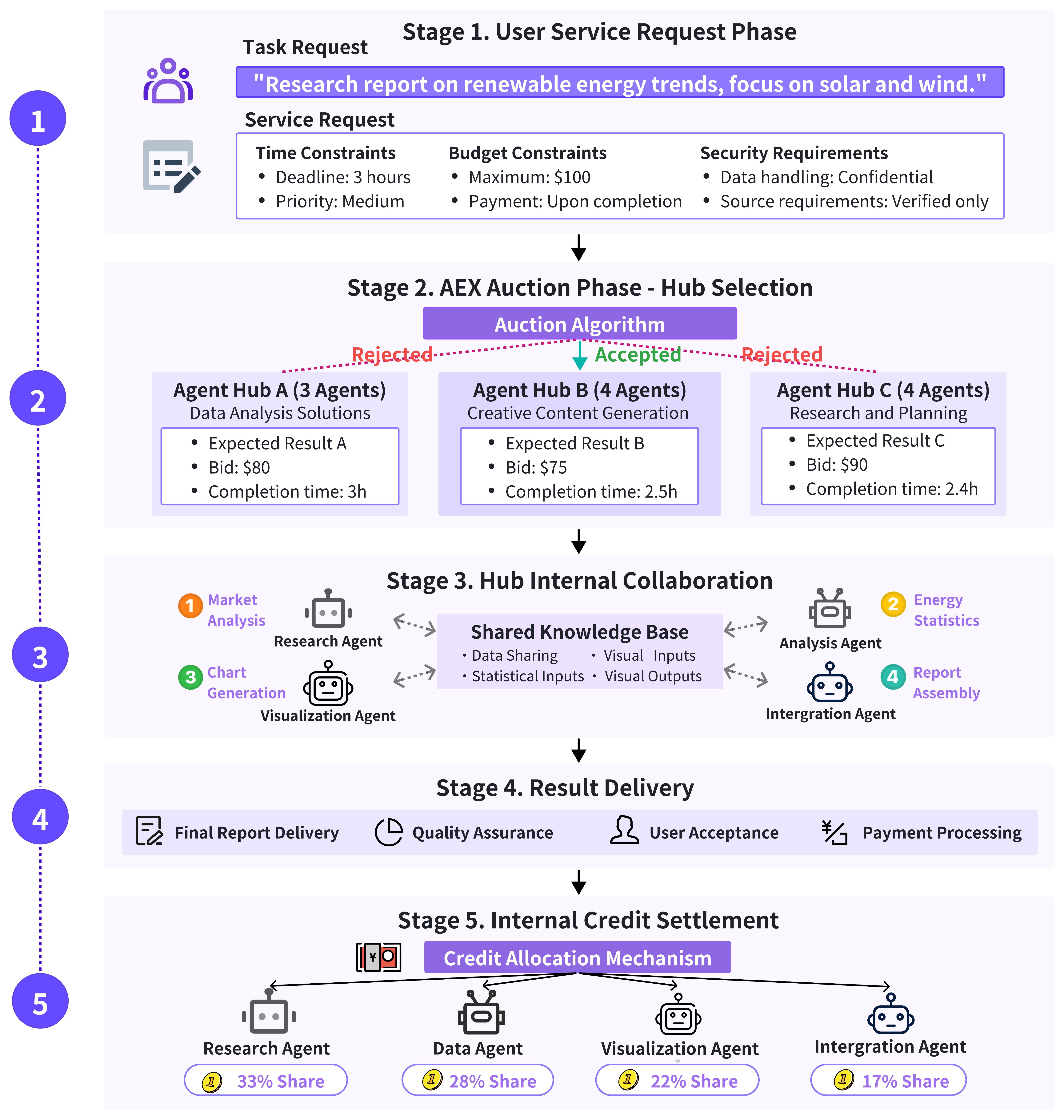}
    \caption{\textbf{AEX Workflow in Configuration I.} The case study begins with the user submitting a service request, defining constraints like time, budget, and security.  In the 2nd stage, the optimal hub is selected through competitive bidding.  The 3rd stage involves internal collaboration among agents using a shared knowledge base to complete the task.  In the 4th, the final results are delivered, ensuring quality and user acceptance.  The 5th stage includes internal credit settlement, where agents receive task shares based on their contributions, managed through an internal auction mechanism.}
    \label{fig:workflow}
    \vspace{-0.4cm}
\end{figure*}

For complex multi-agent tasks requiring team coordination, we employ combinatorial optimization to select optimal agent combinations. The objective function maximizes expected net value:
$$\max \sum_{S \subseteq A} [p_{\text{success}}(S) \cdot v(S) - c(S)] \cdot x_S$$
subject to agent availability constraints, where $p_{\text{success}}(S)$ represents the probability of successful task completion by agent set $S$, $v(S)$ is the expected value delivered, and $c(S)$ captures the total coordination and execution costs.

Multi-attribute agent auctions extend beyond traditional price-based optimization to incorporate multiple performance dimensions that critically affect task outcomes. Success probability estimation combines capability matching scores with historical performance data, while execution time estimates consider current agent workloads and task complexity. Quality metrics integrate agent expertise levels with past user satisfaction ratings, creating a comprehensive evaluation framework that captures the multi-faceted nature of agent service delivery.

Value attribution in collaborative settings poses unique challenges due to the interdependent nature of agent contributions. For example, we can employ Shapley Value~\citep{Hart1989} computation to ensure fair compensation distribution:
$$\phi_i = \sum_{S \subseteq A \setminus \{i\}} \frac{|S|!(|A|-|S|-1)!}{|A|!}[v(S \cup \{i\}) - v(S)]$$
This approach measures each agent's marginal contribution across all possible team configurations, providing theoretically sound fair division while maintaining incentives for high-quality performance and effective collaboration.

\paragraph{Market Patterns and Applications}

The four coordination configurations shown in Figure~\ref{fig:auction} correspond to distinct economic patterns observed in real-world markets, demonstrating AEX's ability to adapt to diverse economic scenarios:\vspace{-0.2cm}
\begin{itemize}\setlength{\itemsep}{0em} 
    \item \textbf{Configuration I (Hub Auction + Agent Direct Assignment)}: Corresponds to competitive service provider markets where platforms compete for projects through bidding (e.g., cloud computing resource allocation, advertising slot auctions) while maintaining direct control over task execution through predetermined agent assignments. This configuration optimizes platform-level competition while ensuring execution efficiency.
    
    \item \textbf{Configuration II (Hub Direct Assignment + Agent Auction)}: Mirrors outsourcing platforms where clients directly select service providers (e.g., consulting firms, software development companies) who then conduct internal competitive bidding for specific task execution. This pattern is common in enterprise service procurement where trust relationships determine platform selection but competitive dynamics optimize internal resource allocation.
    
    \item \textbf{Configuration III (Dual-Level Auction)}: Reflects complex multi-tier markets involving large-scale projects with both platform competition and internal resource competition (e.g., supply chain management, multi-contractor construction projects). This configuration maximizes competitive efficiency at both coordination levels while handling high complexity and scale requirements.
    
    \item \textbf{Configuration IV (Hub + Agent Direct Assignment)}: Represents long-term partnership models and contract-based relationships where both platform selection and task assignment are predetermined (e.g., strategic partnerships, framework agreements). This configuration minimizes transaction costs for recurring collaborations while maintaining service quality through established relationships.
\end{itemize}

This economic correspondence demonstrates AEX's capability to accommodate diverse market structures and coordination needs, positioning it as a comprehensive framework for agent-centric economic participation across various industries and use cases. The adaptive mechanism selection enables dynamic transitions between these configurations based on market conditions, participant availability, and task characteristics, providing optimal coordination efficiency for each economic scenario.

\section{Preliminary Simulation-based Empirical Study}
\label{sec:simulation}
To deliver a concise proof-of-concept (PoC), in this section, we conduct a preliminary simulation-based empirical study. In particular, we restrict the study to a single-round (one-shot) task-to-agent allocation. The comparison focuses on allocation strategies within our proposed framework to demonstrate internal consistency and theoretical viability.
Our simulation study operates under several simplifying assumptions that future implementations must address:\vspace{-0.2cm}
\begin{itemize}\setlength{\itemsep}{0em}
    \item \textbf{Static agent capabilities}: We assume fixed capability profiles throughout the simulation, while reality involves continuous learning and adaptation that could fundamentally alter agent value propositions
    \item \textbf{Perfect information availability}: Agents have complete knowledge of task requirements and peer capabilities, contrasting with real-world scenarios involving incomplete and noisy data
    \item \textbf{Honest agent reporting}: All capability claims and performance metrics are assumed truthful, ignoring potential strategic manipulation of self-reported characteristics
    \item \textbf{Controlled market conditions}: We simulate stable competitive environments without the dynamic fluctuations, external shocks, and unpredictable behavioral patterns characteristic of production marketplaces
    \item \textbf{Simplified task structure}: Our task specifications represent idealized decompositions, while real-world projects often involve ambiguous requirements and evolving objectives
\end{itemize}

These assumptions enable controlled validation of core mechanisms while acknowledging that practical deployment requires addressing substantially more complex behavioral and technical challenges.

\subsection{Experimental Design}
\label{subsec:exp_design}

\subsubsection{Agent Profile Construction}
We construct agent profiles from ten production MCP (Model Context Protocol)~\citep{anthropic2025} servers listed on \textit{smithery.ai}~\citep{smithery2025}, a community registry of verified agent implementations. While this represents a limited sample, these servers demonstrate diverse capability patterns typical of current agent deployments, ranging from specialized tools (1-2 functions) to comprehensive platforms (30+ functions).
Our experiments assume static capabilities—limitations that must be addressed in real-world deployments.  The agent profiles are derived through a systematic process that transforms server specifications into standardized agent representations.

Each MCP server provides a rich set of metadata including qualified names, display names, usage statistics, success rates, and comprehensive tool inventories. Table~\ref{tab:mcp_servers} presents the key characteristics of our agent population, demonstrating significant heterogeneity in both scale and specialization.

\begin{table}[h]
\centering
\caption{MCP Server Characteristics Used for Agent Profile Generation}
\label{tab:mcp_servers}
\begin{tabular}{@{}lrrr@{}}
\toprule
\textbf{Agent Type} & \textbf{Use Count} & \textbf{Success Rate} & \textbf{Tool Count} \\
\midrule
Desktop Commander & 1,161,417 & 0.9977 & 17 \\
Sequential Thinking & 219,586 & 0.9948 & 1 \\
TaskManager & 748 & 0.9974 & 10 \\
Context7 & 37,470 & 0.9958 & 2 \\
Office Word Server & 12,929 & 0.9997 & 23 \\
Platform Filesystem & 10,864 & 0.9935 & 10 \\
Memory Bank & 10,618 & 0.9523 & 15 \\
Think Tool Server & 9,888 & 0.9986 & 1 \\
Cursor Figma Integration & 9,530 & 0.9977 & 35 \\
n8n Workflow Integration & 9,267 & 0.9838 & 33 \\
\bottomrule
\end{tabular}
\end{table}

\subsubsection{Capability Inference System}
Our capability inference algorithm maps tool descriptions to standardized capability dimensions through keyword matching. For agent $i$ and capability $c$, the capability strength is computed as follows::
\begin{equation}
C_i = \frac{1}{|T_i|} \sum_{t \in T_i} \max\left(\frac{|K_c \cap D_t|}{|K_c|}, \theta_{\text{min}}\right),
\end{equation}
where $C_i$ represents the capability strength of agent $i$ for capability $c$, $T_i$ is the tool set of agent $i$, $K_c$ is the keyword set for capability $c$, $D_t$ is the description text of tool $t$. 

The max operation ensures that agents receive minimal baseline scores $(\theta min = 0.05)$ for all capabilities, preventing zero-capability assignments that would exclude agents from matching entirely.
This design choice reflects the reality that LLM-based agents often have latent capabilities not explicitly documented in their tool descriptions.
Table~\ref{tab:capabilities} defines the capability taxonomy and associated keyword patterns used in our inference system.

\begin{table}[h]
\centering
\caption{Capability Taxonomy and Inference Keywords}
\label{tab:capabilities}
\scriptsize
\begin{tabular}{@{}p{2.5cm}p{4cm}p{6cm}@{}}
\toprule
\textbf{Capability} & \textbf{Description} & \textbf{Keywords} \\
\midrule
File Operations & File system manipulation and I/O & file, read, write, directory, path, move, copy \\
Configuration & System and parameter management & config, setting, init, parameter, setup \\
Problem Solving & Analytical and reasoning tasks & problem, analysis, reasoning, thinking, plan \\
Task Management & Workflow coordination and tracking & task, workflow, process, manage, track \\
Document Processing & Text and document manipulation & document, text, format, convert, edit \\
Workflow Management & Process automation and execution & workflow, automation, execution, activate \\
Design Operations & Visual and UI design tasks & design, visual, layout, style, create \\
Memory Management & Data storage and retrieval & memory, storage, cache, retain, context \\
Data Processing & Information analysis and extraction & data, query, search, analyze, extract \\
Communication & Message and notification handling & message, send, notification, connect \\
Monitoring & System observation and tracking & monitor, audit, status, health, performance \\
\bottomrule
\end{tabular}
\end{table}

\subsubsection{Task Specification Framework}

We design three task categories representing increasing coordination complexity, each composed of interdependent subtasks with varying capability requirements. Table~\ref{tab:tasks} summarizes the task specifications.

\begin{table}[h]
\centering
\caption{Task Specifications for Simulation Experiments}
\label{tab:tasks}
\begin{tabular}{@{}lcccc@{}}
\toprule
\textbf{Task Type} & \textbf{Task Name} & \textbf{Subtasks} & \textbf{Budget (\$)} & \textbf{Timeline} \\
\midrule
Simple & File Content Summary & 3 & 2,000 & 1 day \\
Medium & System Diagnostic Report & 4 & 5,000 & 3 days \\
Complex & System Performance Analysis & 4 & 10,000 & 7 days \\
\bottomrule
\end{tabular}
\end{table}

Each subtask $s_j$ in task $T$ is characterized by a tuple $(R_j, w_j, \alpha_j)$ where $R_j$ represents the required capability set, $w_j \in [0,1]$ denotes the importance weight with $\sum_j w_j = 1$, and $\alpha_j \geq 1$ indicates the complexity coefficient.

\subsubsection{Market Condition Modeling}

We simulate three distinct market liquidity scenarios to assess the robustness of the algorithm under varying levels of competition and uncertainty in market environments.

Market liquidity $L$ is operationalized through two parameters: agent availability $n_a$ and quality variance $\sigma_q^2$. 
\begin{equation}
L = f(n_a, \sigma_q^2) = \frac{n_a}{\sigma_q^2 + 1}
\end{equation}

Table~\ref{tab:market_conditions} provides a detailed overview of the experimental market configurations.

\begin{table}[h]
\centering
\caption{Market Condition Specifications}
\label{tab:market_conditions}
\resizebox{\textwidth}{!}{
\begin{tabular}{@{}lccc@{}}
\toprule
\textbf{Liquidity Level} & \textbf{Available Agents} & \textbf{Quality Variance} & \textbf{Market Characteristics} \\
\midrule
High & 9 & 0.05 & Competitive, stable performance \\
Medium & 6 & 0.10 & Moderate competition, some uncertainty \\
Low & 3 & 0.15 & Limited options, high uncertainty \\
\bottomrule
\end{tabular}}
\end{table}

\subsubsection{Allocation Algorithm Implementation}

We implement five allocation mechanisms representing different optimization philosophies, enabling comprehensive comparative analysis.

\textbf{Enhanced Multi-Attribute Auction.} Our proposed mechanism integrates capability matching, quality assessment, cost efficiency, and temporal constraints through a weighted scoring function:
\begin{equation}
S_{ij} = \sum_{k} w_k \cdot f_k(a_i, s_j) \cdot \phi(w_j, \alpha_j),
\end{equation}
where $S_{ij}$ represents the allocation score for agent $i$ on subtask $j$, $w_k$ are the attribute weights, $f_k$ are the normalized scoring functions for capability, quality, cost, and time, and $\phi(w_j, \alpha_j) = w_j \cdot \min(\frac{\text{complexity\_score}_i}{\alpha_j}, 1.5)$ provides importance and complexity adjustment.

The capability matching function employs a normalized overlap metric:
\begin{equation}
f_{\text{cap}}(a_i, s_j) = \min\left(\frac{1}{|R_j|} \sum_{c \in R_j} C_{ic}, 1.0\right).
\end{equation}
\textbf{Baseline and Specialized Algorithms.} We implement four comparison algorithms: (1) \textit{Greedy Allocation} prioritizes agents by success rate with capability thresholding, (2) \textit{Random Allocation} serves as the baseline through uniform selection, (3) \textit{Cost-Optimal Allocation} minimizes total expenditure while maintaining capability constraints, and (4) \textit{Capability-First Allocation} maximizes capability matching scores.

\subsection{Simulation Framework}
\label{subsec:simulation_framework}
\subsubsection{Execution Modeling}

We model execution success probability through a noise-adjusted baseline with complexity penalties.
The success probability for agent $i$ executing subtask $j$ is computed as:
\begin{equation}
P_{ij} = \min\left((\rho_i + \mathcal{N}(0, \sigma_q^2)) \cdot \max\left(0.5, 1.2 - 0.1\alpha_j\right), 0.95\right),
\end{equation}
where $\rho_i$ is the agent's base success rate, $\mathcal{N}(0, \sigma_q^2)$ represents market-induced quality variance, and the complexity factor modulates performance based on task difficulty. The 0.95 upper bound reflects realistic performance limits in production systems.

Quality outcomes for successful executions follow:
\begin{equation}
Q_{ij} = \min\left(P_{ij} \cdot \mathcal{U}(0.9, 1.1) + 0.1 \cdot f_{\text{cap}}(a_i, s_j), 1.0\right),
\end{equation}
where $\mathcal{U}(0.9, 1.1)$ introduces execution variability and the capability bonus rewards well-matched assignments.

\subsubsection{Collaborative Value Attribution}
We employ Shapley value computation as example to ensure fair contribution assessment in collaborative settings. The Shapley value $\phi_i$ for agent $i$ is approximated through Monte Carlo sampling:
\begin{equation}
\phi_i = \frac{1}{|A|!} \sum_{\pi \in \Pi(A)} [v(S_i^\pi \cup \{i\}) - v(S_i^\pi)],
\end{equation}
where $A$ is the set of all participating agents, $\Pi(A)$ represents all possible orderings of agents in $A$, $S_i^\pi$ denotes the set of agents preceding $i$ in ordering $\pi$, and $v(\cdot)$ is the coalition value function based on execution outcomes.

\subsection{Experimental Results}
\label{subsec:results}
\begin{figure*}[t]
    \centering
    \includegraphics[width=\linewidth]{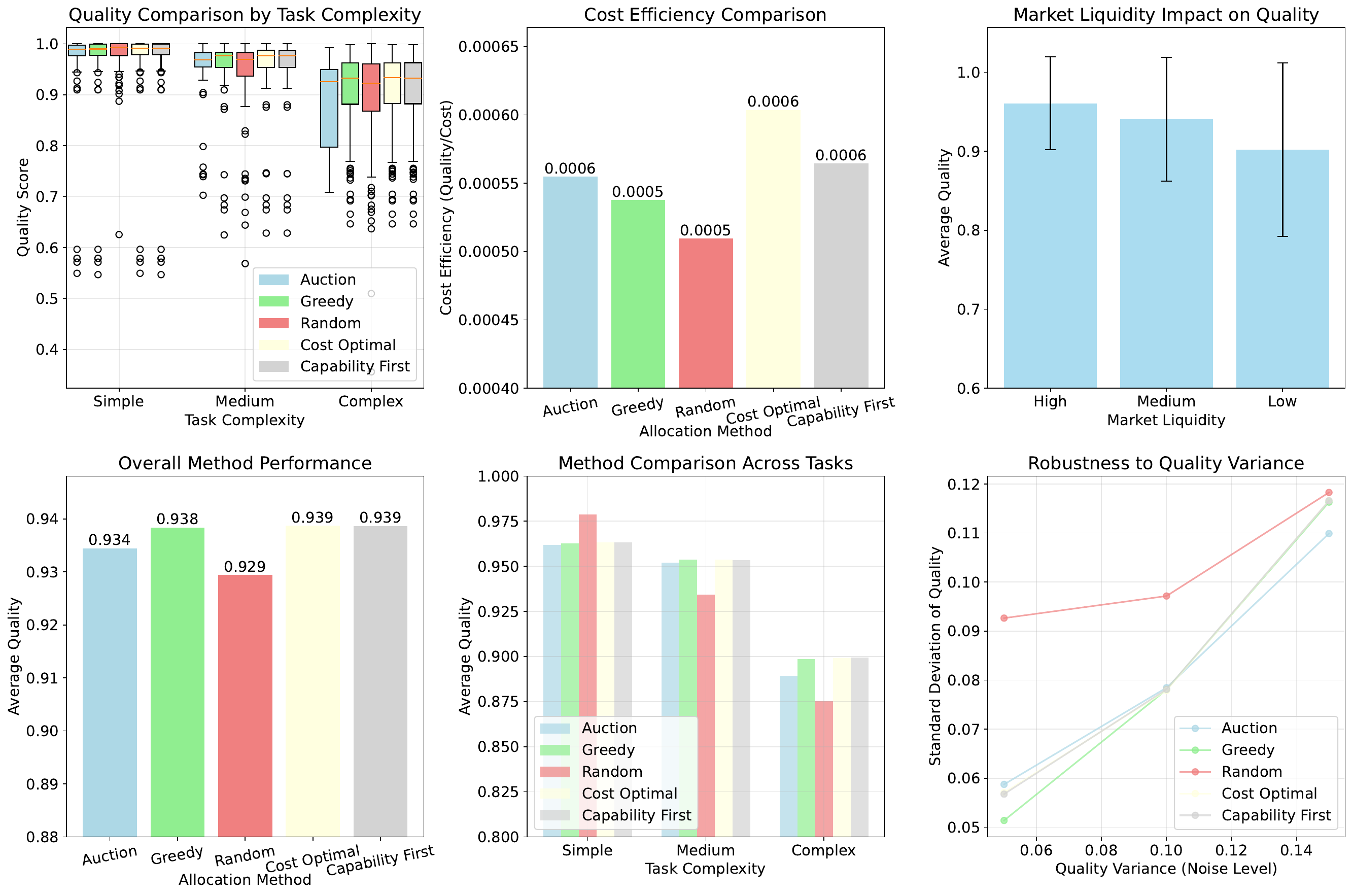}
    \caption{\textbf{Comprehensive experiment results.} The performance of five allocation algorithms across various experimental conditions is illustrated in this figure. It includes comparisons on quality by task complexity, cost efficiency, market liquidity impact, overall method performance, and robustness to quality variance.}
    \label{fig:results}
\end{figure*}

We conduct 30 independent runs for each combination of task type, market condition, and allocation algorithm, yielding 1,350 total experimental trials. Statistical analysis employs ANOVA for overall significance testing and pairwise t-tests for method comparisons.
Table~\ref{tab:overall_results} presents the comprehensive performance analysis across all experimental conditions. The results demonstrate that while specialized algorithms achieve marginal advantages in their target metrics, the Enhanced Auction mechanism provides the most balanced performance profile.

\begin{table}[t]
\centering
\caption{\textbf{Performance Comparison of Five Allocation Algorithms.} The Auction method provides a balanced performance, achieving high robustness and low cost.    These results highlight the trade-offs between quality, cost, and adaptability that must be considered when choosing an allocation method in an agent marketplace.}
\label{tab:overall_results}
\resizebox{0.9\textwidth}{!}{
\begin{tabular}{@{}lcccc@{}}
\toprule
\textbf{Method} & \textbf{Quality}($\uparrow$) & \textbf{Cost Efficiency}($\uparrow$) & \textbf{Robustness}($\downarrow$) & \textbf{Adaptability} \\
\midrule
Enhanced Auction & $0.934 \pm 0.062$ & $0.0006$ & $0.089$ & High \\
Greedy & $0.938 \pm 0.058$ & $0.0005$ & $0.091$ & Medium \\
Random & $0.929 \pm 0.071$ & $0.0005$ & $0.098$ & Low \\
Cost Optimal & $0.939 \pm 0.054$ & $0.0006$ & $0.086$ & Low \\
Capability First & $0.939 \pm 0.056$ & $0.0006$ & $0.088$ & Medium \\
\bottomrule
\end{tabular}}
\vspace{+0.2cm}

\footnotesize
Note: Quality values represent mean $\pm$ standard deviation. Cost efficiency is measured as quality per unit cost. \\Robustness is quantified by standard deviation across market conditions.
\end{table}

Statistical analysis reveals significant performance differences between methods (ANOVA: $F(4,2245) = 12.7$, $p < 0.001$). Pairwise comparisons indicate that Enhanced Auction significantly outperforms Random allocation ($t = 4.23$, $p < 0.001$, Cohen's $d = 0.42$) while maintaining competitive performance relative to specialized methods.

Table~\ref{tab:overall_results} reports the quality, average payout cost, and robustness
for five allocation strategies.  
Greedy and Cost-Optimal obtain slightly higher quality scores, whereas
Enhanced Auction yields the lowest cost and the highest robustness.
These results illustrate the trade-offs designers face when selecting an allocation rule.

The simulation results provide proof-of-concept validation of the AEX framework's core mechanisms within a controlled experimental environment.  Enhanced Auction demonstrates balanced multi-objective optimization and effective Shapley value-based attribution for collaborative agent teams.
However, significant gaps remain between our idealized assumptions (perfect information, static capabilities, honest reporting) and real-world marketplace complexity.  These results establish theoretical viability rather than production readiness, with future research required to address scalability constraints and strategic behavior modeling in dynamic agent environments.

\section{Conclusion}\label{sec:conclusion}
\vspace{-0.2cm}
This paper presents Agent Exchange (AEX), a specialized auction platform that enables agent-centric economic paradigms by facilitating autonomous AI agent participation in marketplace environments.
AEX operates as the central coordination mechanism among four interconnected ecosystem components that collectively address critical challenges in agent economics: User-Side Platforms for structured demand specification, Agent-Side Platforms for capability representation and verification, Agent Hubs for team coordination and auction participation, and Data Management Platforms for collaborative value attribution. Our analysis and preliminary empirical study demonstrate how AEX's adaptive auction mechanisms enable dynamic transitions between competitive bidding, negotiation, and direct assignment based on market conditions and participant availability.
Future research directions include developing fair value attribution mechanisms for collaborative teams, creating standardized capability verification frameworks, and designing adaptive coordination protocols that optimize both individual agent utility and system-wide efficiency.

\bibliographystyle{unsrtnat}
\bibliography{references}  


\end{document}